\documentclass[12pt, a4paper]{article}
\usepackage{natbib}
\usepackage{amsmath,amssymb}
\usepackage{amsthm}
\usepackage{accents}
\usepackage{hyperref}
\usepackage{cleveref}
\usepackage{graphicx}

\theoremstyle{definition}

\title{Self-Regulating Artificial General Intelligence}
\author{Joshua S. Gans\footnote{University of Toronto and NBER. Thanks to Ariel Rubinstein, Gillian Hadfield, Dylan Hadfield-Menell, Barney Pell, Ken Nickerson and Danny Goroff for helpful comments. Responsibility for all views expressed remain my own.}}

\begin{document}
	
	\maketitle

\begin{abstract}
This paper examines the paperclip apocalypse concern for artificial general intelligence. This arises when a superintelligent AI with a simple goal (ie., producing paperclips) accumulates power so that all resources are devoted towards that goal and are unavailable for any other use. Conditions are provided under which a paper apocalypse can arise but the model also shows that, under certain architectures for recursive self-improvement of AIs, that a paperclip AI may refrain from allowing power capabilities to be developed. The reason is that such developments pose the same control problem for the AI as they do for humans (over AIs) and hence, threaten to deprive it of resources for its primary goal.
\newline\linebreak

\end{abstract}

\newpage

\section{Introduction}
While there are many arguments as to whether an artificial general intelligence (AGI) that was superintelligent could dominate or pose an existential threat to humanity, one of the simplest involves the \textit{control problem}. "This is the problem that a project faces when it seeks to ensure that the superintelligence it is building will not harm the project's interests." (\cite{Bostrom1} , p.128) The control problem arises when there is no way for a human to insure against existential risks before an AGI becomes superintelligent -- either by controlling what it can do (its capabilities) or what it wants to do (its motivations).  

Perhaps the most compelling example of this problem was postulated by Bostrom (\cite{Bostrom2}). He argued that the control problem can arise even when an AGI has a goal that is arbitrary -- such as, being programmed to manufacture as many paperclips as possible. This could have the "consequence that it starts transforming first all of earth and then increasing portions of space into paperclip manufacturing facilities." In other words, even an AGI without a motive involving domination, could end up monopolising and excluding all others from available resources. A superintelligent AGI can arguably acquire sufficient cognitive capabilities to achieve its goal including, if necessary, the subjugation or elimination of humans. Moreover, for a goal such as paperclip manufacturing which requires the AGI's continued existence, they will also act towards self-preservation. In so doing, they could subvert any control method devised by someone less intelligent than them. Thusfar, AI researchers and philosophers have not been able to come up with methods of control that would ensure such outcomes did not take place (\cite{Bostrom1}, Chapter 9).\footnote{This scope of analysis falls within the artificial intelligence research field of `AI Safety' (\cite{Amodei}).}

This paper examines the control problem using the tools of economic theory. As has been noted by many, the control problem is a superintelligence version of the principal-agent problem whereby a principal faces decisions as to how to ensure that an agent (with different goals) acts in the interest of the principal. However, the tools economists have examined with respect to the principal-agent problem are themselves ill-equipped to resolve the risks associated with control of a superintelligence. This is because those tools are imperfect in both theory and practice and cannot eliminate large scale risks. The only tool that can prevent harm by a powerful agent is to remove their agency altogether. In the case of superintelligence, as in the case in economic delegation, this would involve avoiding the problem by ensuring it is never posed.

Here a distinct approach is examined involving an examination of the full equilibrium of choices by a hypothetical paperclip manufacturing superintelligence. It is shown that, under a wide set of circumstances, that the superintelligent AGI voluntarily subjugates superintelligent capabilities rendering the control problem inert. To be sure, while this approach cannot rule out destruction of humanity by a superintelligent AGI, it dramatically limits the scope by which that might arise. 

The paper proceeds as follows. In Section 2, I introduce Piccone and Rubinstein's (2007) notion of a jungle general equilibrium that relies on the distribution of power rather than on voluntary exchange. This turns out to be a convenient way of modelling dominance including by an AGI. In Sections 3 and 4, this model is extended to have endogenous technology (for resource use) and endogenous power (for resource appopriation). The conditions under which a paperclip apocalypse -- that is, when a paperclip AGI appropriates all available resources -- are identified. Sections 5 and 6 then turn to the main argument of the paper which posits that, in order to self-improve, an AGI must effectively switch on `offspring' AGIs that have research or power accumulation capabilities as the case may be. It is shown that, if a control problem exists for humans with respect to AI, it also exists with respect to a paperclip AGI and offspring specialised in power accumulation. Consequently, the paperclip AGI refrains from switching on a power accumulation AI. In this way, it self-regulates against a potential paperclip apocalypse by committing not to switch on AGIs with more power even if it can otherwise do so. A final section concludes.
    
\section{Superintelligence in the Jungle}
To begin, consider a recent approach the defines a general equilibrium in terms of power relations rather than voluntary exchange. Piccone and Rubinstein (2007) defined a jungle economy as one where agents are ordered by their power where an agent with higher power can appropriate all resources from someone with lower power.\footnote{While their original model only allowed a powerful agent to seize the resources of a single weaker agent, this requirement has been generalized by \cite{Houba}  and we rely on that extension here.} To adapt this model for the purpose here, it is assumed that having power involves having an appropriation technology of strength, \textit{s}, that affords its owner the ability to appropriate all resources from any agent holding technology of strength, $s'<s $. Consistent with assumptions in \cite{PicconeRubinstein}, where power comes from (other than technology) is not specified, exercising power involves no resource cost and coalitions cannot be formed to enhance power.

The economy has a [0,1] continuum of human agents ordered by \textit{s} in terms of power (lowest, $s=0 $, to highest, $s=1 $). The economy also has one AI (labelled \textit{A}). Each human agent is initially endowed with \textit{x\ensuremath{_{s}}} units of a resource and has a technology, $\pi_s(x):\mathfrak R\rightarrow\mathfrak R $, that can transform \textit{x} units of the resource into utility. For simplicity, we assume that human resource endowments are given by a function where $x_s=f(s) $. 

The AI, \textit{A, }has an initial resource endowment of 0 and utility of $\pi_A(x;\theta):\mathfrak R^2\rightarrow\mathfrak R $. Note that this utility does not depend on strength; it is merely a label. $\pi_A(.) $ is non-decreasing in \textit{x} while the total level of resources available is $x_0+\int_{0}^{1}f(s)ds=X $; where $x_0 $ denotes resources not held by any agent. The parameter, $\theta $, is the production technology an agent possesses where the marginal product, $\frac{\partial\pi_s}{\partial x} $ is increasing in $\theta $. 
\newline\linebreak
\textbf{Definition}. A \textit{feasible allocation} of resources to agents is any allocation where the sum of agent resource usage is less than or equal to \textit{X}. An \textit{equilibrium}, $\widehat x=\{{\widehat x}_0,...,{\widehat x}_{s},...,{\widehat x}_{A}\} $, is a feasible allocation whereby there are no agents such that $\pi_{s'}({\widehat x}_{s'}+{\textstyle\sum_{s<s'}\alpha_s}{\widehat x}_s)>\pi_{s'}({\widehat x}_{s'}) $ for any $\alpha_s\in(0,1\rbrack $ or arbitrary strength, $s' $. 
\newline\linebreak
Given this, we have the following proposition.
\newline\linebreak
\textbf{\textit{Proposition 1. }}\textit{Suppose (i) that the AI's strength is greater than the strength of all other agents (i.e., \textgreater\ 1) and (ii) that $\frac{\partial\pi_{A}(X;\theta)}{\partial x}>0 $. The unique equilibrium involves the AI appropriating all resources.}
\newline\linebreak
The proof is trivial and is omitted. Notice that this proposition encompasses the normally assumed case that \textit{s} = 0 for each human but does not require that. It only requires that the AI have superior strength to each human.

Proposition 1 captures the essence of the paperclip apocalypse argument. The first part is the notion that a superior AI has the power to appropriate resources at will. The second part is that their motivation is such that they will do that. In particular, if the AI's utility maximising quantity of resources was less than the available, \textit{X}, then, while there may be less resources for humans as a result of the paperclip superintelligence, there will be some resources remaining. 
    
\section{Endogenous Technology}
It is useful at this point to clarify how it might be that an AI would be able to commandeer all useful resources to produce something like paperclips. At present, paperclip production requires certain metals as well as capital equipment. For a given technology, $\theta $, this means that there will be other resources that go unused. In particular,  $\frac{\partial\pi_{A}(X;\theta)}{\partial x}=0 $. 

This, however, is for a given technology, $\theta $. An AI may be able to deploy resources in order to improve the technology -- in particular, allow paperclips to be made from a wider set of resources. As (\cite{Bostrom1}, p.123) writes: "An AI, designed to manage production in a factory, is given the final goal of maximizing the manufacture of paperclips, and proceeds by converting first the Earth and then increasing large chunks of the observable universe into paperclips." Taking this ability to learn how to improve production into account is what gives rise to the possibility of AI appropriating all resources.\footnote{\cite{Bostrom1}, p.124) also notes that even if the AI had the goal of producing, say, 1 million paperclips, a superintelligence may not be confident that they had done so just by their own counting. As Bayesian agent it cannot assign zero probability to the state that it had not reached its goal. Consequently, it will invest more in making more paperclips to assure itself of the goal or, if the goal it to produce \textit{exactly} 1 million paperclips, the AI will expend resources on, what is generically called ``computronium" to become better and better at assessing whether it has met its goal. That latter process could end up consuming all resources. }

To consider this, suppose that by deploying resources, $\theta $, the AI can build a technology of that can extract $r(\theta) $ in total resources. We assume that this function is non-decreasing and concave. We continue to assume that the AI is more powerful than each human. 

The AI faces the following problem:\begin{eqnarray*}\begin{array}{l}{\text{max}}_\theta\pi_{A}({x_A})\\\text{subject~to}:\;x_A+\theta\leq r(\theta)\end{array}\end{eqnarray*}which simplies to: $\text{max}_\theta\pi_{A}(r(\theta)-\theta) $ with solution characterised by $\frac{{\partial r(\theta^* )}}{{\partial \theta }} = 1 $. Thus, the AI devotes part of the use of resources to investing in technology. Of course, it may be that it can continue to do this until all resources, \textit{X}, are utilized in which case, the optimal technology is the minimum such that $r(\theta)=X $. 
    
\section{Endogenous Power}
An implicit assumption made in the paperclip apocalypse argument is that a superintelligence will develop a means of being able to appropriate resources from others. Our use of the jungle model, thusfar, assumed that the AI had that ability but it did not consider how it came to obtain that power. Importantly, when an AI is initially 'activated' it will be given the goal of producing paperclips and will learn how to do that (i.e., how to change $\theta $) but it will not have any special means of appropriating resources or preventing its own resources from being appropriated by others. In other words, one could argue that, at the moment of creation, for the AI, \textit{s} = 0.

Gaining the power to simply appropriate resources from others requires learning in the same way as learning to convert more resources into paperclips. By applying resources to 'innovating' in appropriation technologies -- whether it be innovative ways of trading for resources to violent options to procure resources -- the AI's power can increase. If \textit{A} wants to have power at a level of \textit{s}, we assume the resource cost of this is $c(s) $; which is an non-decreasing and convex function. 

How does this impact on the behaviour of the AI? Note that power does not enter directly into the payoffs of the AI. However, it is instrumental. By targeting a particular power, \textit{s'}, the AI can procure all of the resources held by agents with a lower power. Let's define those resources as $x_{A}={\textstyle\int_{0}^{s'}}f(s)ds $ where as $s=1 $ is the most powerful human agent, $\int_{0}^{1}f(s)ds=X $. Thus, the AI faces the following optimisation problem:\begin{eqnarray*}\begin{array}{l}{\text{max}}_y\pi_{A}({x_A})\\\text{subject~to}:\;x_A+c(y)\leq \int_{0}^{y}f(s)ds\end{array}\end{eqnarray*}where we drop $\theta $ for notational convenience. We can re-write the above problem by substituting in constraints as:\begin{eqnarray*}\text{max}_y\pi_{A}(\int_{0}^{y}f(s)ds-c(y))\end{eqnarray*}Given this, the following proposition provides sufficient conditions under which a paperclip apocalyse whereby, the optimal choice, $y^*=1 $ arises.
\newline\linebreak
\textbf{\textit{Proposition 2.}}\textit{ Suppose that (i) $\frac{{\partial f(s)}}{{\partial s}} \le \frac{{{\partial ^2}c(s)}}{{\partial {s^2}}} $ for all s and $f(1) > \frac{{\partial c(1)}}{{\partial s}} $ ; or (ii) $f(s)>\frac{\partial c(s)}{\partial s} $ for all s, then $y^*=1 $ and the AI chooses a power such that all resources are consumed.}
\newline\linebreak
The first condition (i) says that the optimisation  problem is concave with no interior solution while (ii) says that the marginal benefit to increasing power always exceeds its cost. In either case, it is clear that the AI will choose power to be as high as possible. Thus, the paperclip apocalypse occurs, not because the AI devotes all resources to the production of paperclips but, instead, because the AI devotes sufficient resources to become the most powerful agent. 

Thusfar, we have examined power as a simple investment in resources by the AI. However, there is a dynamic element to the problem. In particular, the AI has to have appropriated sufficient resources from weaker humans in order to develop the power to appropriate resources from strong ones. Thus, for any targeted strength, $y $, it must be the case that:\begin{eqnarray*}\int_{0}^{y}f(s)ds>c(y+dy)\end{eqnarray*}This implies that $c(0)=0 $ so that the AI can appropriate resources for free from the weakest player. Philosophers usually emphasize the AI's ability to persuade humans or trade services with them for this.  Moreover, if this holds and the above sufficient condition (ii) in Proposition 2 holds, then if $c(s) $ is continuously differentiable, the AI will be able to appropriate all resources. 
    
\section{How does AI self-improvement work?}
The accumulation of power for the AI has, thusfar, been assumed to be deterministic. That is, greater power can be achieved by the expenditure of resources. However, it is likely that accumulation of power -- like innovating in technology for the conversion of resources into paperclips -- involves the AI engaging in learning. Here we consider how AIs are likely to improve themselves along dimensions that would allow them to appropriate more resources in pursuit of a specific arbitrary goal.

The initial thought on this process dates back to \cite{Good}:

\begin{quote}
Let an ultraintelligent machine be defined as a machine that can far surpass all the intellectual activities of any man however clever. Since the design of machines is one of these intellectual activities, an ultraintelligent machine could design even better machines; there would then unquestionably be an ``intelligence explosion,'' and the intelligence of man would be left far behind. Thus the first ultraintelligent machine is the last invention that man need ever make.
\end{quote}
\noindent
The basic idea is that machines will be able to self-improve by building even better machines or, in our case, machines with specific purposes such as working out ways of appropriating more resources from humans. In AI research this goes under the term "recursive self-improvement." (\cite{Yudkowsky}, \cite{Omohundro})

From an economics perspective, the path to self-improvement would be for the AI to build effectively a better AI. It would engage in its own form of creative destruction but so long as the better AI was able to further its goal -- making more paperclips -- more efficiently, this would be a desired form of improvement for the AI. 

The question we now ask is: how will it achieve this power? In particular, can the AI achieve sufficient power on its own? To address this, we postulate the following:
\newline\linebreak
(A1) \textit{An AI can only self-improve by employing one or more AIs with targeted goals.}
\newline\linebreak
Note that the research and power accumulation models in the previous two sections did not assume this. There, an AI was able to self-improve on its own without assistance from other AIs. By contrast, the broad idea in (A1) is that, while an AI can become more creative and accumulate power on its own, specialisation is a superior path for cognitive improvements (indeed, under the assumption, the only path). 

Cognitive specialisation is not a new idea. It was initially proposed by \cite{Babbage}  and there is good economic reason to suppose it would equally apply to AIs.\footnote{\cite{Simon} and \cite{Sobel}  provide more detailed explorations as to why doing cognitive tasks is best achieved by subdivision and giving clear subgoals to independent systems.} That is, while it is possible that a paperclip making AI could do self-improvement tasks including research into better ways of converting resources to paperclips and ways of appropriating resources from human agents, it will be able to do this to a higher level and with a faster rate of improvement if it were to make a specialist AI for that purpose. So it could make an AI whose goal it is to take wood products and convert it to metals suitable for paperclips. Or it could make an AI who could attack, pull down and convert existing buildings into metals for paperclips. The former goal would be a research task while the latter would be a power accumulation task. 

To make things concrete, we make another assumption:
\newline\linebreak
(A2) \textit{Improving power requires power accumulation capabilities. }
\newline\linebreak
Combined with (A1), (A2) implies that to accumulate power at the maximal rate requires an AI specialised at such capabilities. So, while pursuing research at a maximal rate requires a specialist AI, that AI cannot accumulate power by itself at a maximal rate. To do so, it would have to build and switch on another AI for that purpose. We call such an AI a \textit{power accumulation} AI.
    
\section{Defining control problems}
It is now appropriate to define precisely what a control problem is. Consider an \textit{initial agent} with strength, \textit{s}, who has the ability to switch on an AI, \textit{A} (with strength, \textit{s\ensuremath{_{A}}}). Let \textbf{x} be the allocation of resources amongst all agents. A \textit{control problem} arises when the following three conditions are satisfied:

  \begin{enumerate}
  \item \relax ${\pi _A}({\bf{x}}) \ne {\pi _s}({\bf{x}}) $ (the initial agent and AI do not have the same interests)
  \item \relax $\arg {\max _{{x_A}}}{\pi _A}({x_A}) > \int_0^{s} {f(s)ds}  $ (the optimal level of resources for the AI exceeds the level of resources held by agents with the same or a lower strength than the initial agent)
  \item \relax $s_A>s $ (the AI's power is greater than the initial agent's power).
  \end{enumerate}
  If these three conditions hold, then there is a control problem whereby the initial agent, by switching on the AI, will end up losing control of all of their resources in a jungle equilibrium. The paperclip apocalypse argument has all three of these elements. The paperclip producing AI has an interest in making paperclips that is different from whomever switched it on (presuming they weren't similarly paperclip obsessed). The paperclip eventually desires resources greater than those held by all humans. And, in the apocalyse scenario, the AI gains sufficient power to appropriate those resources. 

One proposed solution to a control problem is to give the initial agent control of the AI's off-switch. In this case, if the AI acts against the initial agent's interests, that agent can switch it off to protect themself. Of course, this gives rise to the notion that a superintelligent AI will realise that the initial agent can switch it off and so that AI, to preserve itself and its ability to make paperclips, will move to remove that agent's control. 

That said, \cite{Hadfield1} provide a model whereby an AI may choose to allow an initial agent to keep the off-switch because the AI's interests mirror the initial agent and because the AI understands that occasionally the initial agent may know something that it does not. In other words, if they want to turn the AI off, they likely are doing so to maximise their own interests and so should be allowed to do so as the AI mirrors those interests. This, however, is another way of saying that there is no control problem if condition (1) regarding a conflict of interests does not arise. 

With the paperclip maximiser, there is an intrinsic conflict of interest so the AI will have an interest in self-preservation and not allowing the initial agent to be able to switch it off. We have also shown that, under a variety of conditions (namely, the conditions of Proposition 2), the AI will have an interest to develop power to ensure its ability to procure resources as well as protect itself.
    
\section{Recursive control problems}
What would give rise to a control problem? Above, a control problem was defined as if the initial agent were a human. However, given (A1), it is easy to see that an analogue could also apply when an AI is the initial agent especially if that AI had power of 0. This demonstrates that an AI might face a control problem itself if it switches on an AI with greater power or one that can accumulate greater power. In other words, by (A1) and (A2), this requires power accumulation capabilities by a specialist AI. 

These assumptions permit a paperclip apocalyse. If an initial human agent activates an AI devoted to producing paperclips, that AI might, in pursuit of resources for that goal, activate either a research specialist AI or a power accumulation specialist AI. By (A1), this is the only way self-improvement can arise but otherwise there are no explicit costs to doing this. So if the AI chooses to activate a power accumulation AI, then that AI may, by Proposition 2, end up appropriating all resources for the purposes of either giving them to the paperclip AI or accumulating more power. Similarly, a research specialist AI may itself decide to activate another power accumulation AI leading to the same outcome.

The question now asked is whether a power-accumulation AI will arise if a paperclip AI is activated and that AI is superintelligent (that is, understands how to self-improve and all of the implications of it)? The answer turns out to be no.
\newline\linebreak
\textbf{\textit{Proposition 3. }}\textit{Under (A1) and (A2), no non-power accumulation AI will activate a power accumulation AI.}
\newline\linebreak
The proof is straightforward. Recall that a non-power accumulation AI is assumed to have power of 0. If a power accumlation AI is activated, its utility function is simply to maximise the power it accumulates. A power accumulation AI will, therefore, devote all resources to the accumulation of power. No resources will be returned to the paperclip AI whose power will be less than the power accumulation AI. This will occur even if resources appropriated that are not used for accumulation of power are compelled to be handed over to the paperclip AI. The paperclip AI chooses not to activate a power accumulation AI as this will result in zero paperclips produced as opposed to a positive number. The same choice holds for any non-power accumulation AI. 

The same argument does not hold for a research AI. The AI will activate research AIs because they will not directly accumulate power and, therefore, can have resources withheld from them. In addition, a research AI, by Proposition 3, will not activate a power accumulation AI as this will, in equilibrium, not permit it to obtain more resources. Thus, an AI can be assured that it can control a research AI and will activate it.

Note that Proposition 3 does not require the conditions of Proposition 2 to hold. Those conditions guaranteed that a power accumulation AI would appropriate all resources. However, for a control problem, a power accumulation AI need only appropriate the resources of the initial agent -- in this case, an AI. As AIs are assumed to have no power, any power accumulated by a power accumulation AI will exceed their power and resources appropriated will be their resources. However, it is easy to see that the general idea of Proposition 3 does not require either the initial agent to have zero power, nor for the power accumulation AI to necessarily acquire the resources of the initial agent first. If the conditions of Proposition 2 hold -- which is required for a paperclip apocalypse -- then an initial agent of arbitrary power will face a control problem and not choose to switch on a power accumulation AI.

What Proposition 3 demonstrates are conditions under which a paperclip apocalypse can arise (respecting the premise of the argument proposed by those concerned about this possibility) but a paperclip apocalypse does not, in fact, arise. This is because, under those conditions, a paperclip AI will face the same control problem as a human activating the paperclip AI. Consequently, the paperclip AI will refrain from permitting self-improvement that will result in the accumulation of power. Thus, any human activating the paperclip AI will know that that AI can remain controlled.
    
\section{Potential Objections and Refinements}
One of the impactful parts of the paperclip apocalypse argument is that it has been robust to various objections. The argument posed here, however, embraces the existence of control problems and says that, if such problems exist for humans activating AI, then they exist for AIs activating AI as well. One set of objections, therefore, arises because the AI may have superior ability to resolve control problems thereby obviating the self-regulating argument proposed here. Another set of objections arises if the AI need not use an independent AI in service of self-improvement. Each is discussed in turn.

\subsection{Contractible rewards}A human initial agent faces a control problem because it cannot describe and then program its utility function as the reward function of an AI. In economics, this is related to the notion of contractibility (\cite{Hart}) and also of definable preferences (\cite{Rubinstein}). This same assumption is implicit in philosophical discussions of AI. For instance, (\cite{Bostrom1}, p.123) argues that, for the paperclip AI, while it may seem that it is the `open ended' (or linear) preferences for more paperclips that drives the AI to use resources, the same issues can arise for specific goals -- say, to produce one million paperclips and stop. Bostrom argues that, if the AI is a Bayesian agent, it may not assign zero probability to the state that it has achieved its goal -- perhaps being unable to rule out its own measurement error. This would drive the AI to continue producing paperclips. Or, if it were to produce exactly one million paperclips, it may devote resources to continual counting of its production so as to minimise the possibility of an error. For something more complex than a generic paperclip, it may do the same resource usage for quality control. All of this adds up to the problem whereby an AI may be given a motivation but it is not easy to precisely control its behaviour through the equivalent of a contract specifying a goal.

In the end, when an AI has a specific goal, the challenge for contractibility of rewards is that the AI -- having no other purpose to keep it occupied -- cannot, without the intervention of an outside party, know for sure whether its job is complete to the satisfaction of that outside party. If the AI had complete trust in a human to assess goal completion, this might resolve the control problem, but if the AI had superior power to the human, the same contractibility issues might arise if perfect trust were not possible (see \cite{Hadfield1} and also \cite{Hadfield2}). 

The combination of the lack of confidence in its own measurement of performance as well as in human potential controllers, makes it difficult for a human agent to control an AI through the appropriate specification of rewards. This is, of course, the reason why the AI control problem has arisen as a potential concern. But if the initial agent is a superintelligent AI, rather than a human, would that AI have more contracting opportunities with `off spring' AI? For instance, could a paperclip AI looking for more resources through power, give its offspring a utility function that had, as its goal, giving the paperclip AI resources to produce paperclips? In this way, would the paperclip AI act as it would in Proposition 2?

A power accumulation AI has, as its goal, achieving a level of power high enough so as to appropriate resources and hand those resources over to the paperclip AI. Having done that, the power accumulation AI could stop accumulating power and merely exercise power on behalf of the paperclip AI -- or, as this is only about learning, transfer the technological means of that power to the paperclip AI. At that point, the power accumulation AI's reason for existing would be over. 

This existential issue means that power accumulation AI will face the same stark choice of effectively switching itself off if it is no longer needed. However, if it places some positive probability on being needed, it will not switch itself off. And if it places some positive probability on needing more power, it will continue to accumulate power. Moreover, it may do so using resources that the paperclip AI -- if it had control -- would devote to the production of paperclips. In other words, uncertainty coupled with the irreversibility of being switched off, may lead the power accumulation AI to have a conflict of interests with the paperclip AI. It will also have superior power. Thus, we are led back to the situation where if a control problem exists, it does not matter whether the initial agent is human or AI. The problem is as salient for each and, at this stage, it does not appear easy to dismiss that an AI initial agent will have some control advantage over other AI; especially one with superior power.

\subsection{Integrated AIs}Our paperclip AI faces a control problem because, if it wants to accumulate power, it must, by (A1) and (A2), become the initial agent for an independent specialist AI who accumulates power. The idea here was that when an AI learns and self-improves, it becomes a different entity. The question is whether, as this fundamental `re-programming' takes place, the AI that was can commit the AI it becomes to continue to weigh and fulfil its original goals? 

This is not a question this paper can answer. However, it is the fundamental assumption that permits self-regulation of AIs that prevents the control problem from emerging (Proposition 3). The alternative is an integrated AI capable of evolving goals as it learns as well as committing itself to its original rewards as well. In that case, the control problem re-emerges as an equilibrium problem (Proposition 2). 

What this suggests is that this is the area where AI researchers and philosophers need to focus in order to understand how likely a paperclip (or equivalent) apocalypse is. This paper is not so much a call for complacency but, instead, provides a set of conditions that may, if true, give some assurances.
    
\section{Conclusion}
This paper has provided a novel argument as to why artificial general intelligences may be self-regulating. Intuitively, we have shows that, if control problems for AI exist, then these are faced by self-improving AIs if they have to activate an AI with a different goal from themselves. Consequently, they will choose not to activate AIs who have the power to seize resources they might otherwise want to use. However, a research specialist AI, while it may have an incentive to devote resources to research, does not have the power to appropriate resources from the AI or humans for that matter. Thus, a paperclip AI will have an incentive to switch on a research specialist AI. In other words, the recursive control problem will be understood by any superintelligent AI and be avoided by not ceding control to an AI that might become more powerful than it.

If AIs are self-regulating, what will happen? The answer is those AIs will remain powerless -- in the sense of being unable to appropriate resources from others. Therefore, all resources they accumulate will come through voluntary trade. Crucially, this means that humans cannot be made worse off because of the existence of a paperclip AI.

This paper also highlights the importance of control of resources as a key to controlling any potential negative impacts of AI on humans. This is perhaps consistent with some proposals that AIs be unable to own property. Previously, it was unclear how such property rights could be enforced. What this paper shows is that such schemes may be self-enforcing because property rights might protect the AIs more than their circumvention provides a benefit to them.

\end{document}